\pdfoutput=1

\documentclass[11pt]{article}

\usepackage{ACL2023}

\usepackage{times}
\usepackage{latexsym}
\usepackage{amsmath}
\usepackage{amssymb}
\usepackage{graphicx}
\usepackage{tabularx}
\usepackage{multirow}
\usepackage{multicol}
\usepackage{caption}
\usepackage{verbatim}
\usepackage{booktabs}
\usepackage{fontawesome5}
\usepackage{subcaption}
\usepackage[draft]{minted}
\usepackage[noabbrev,capitalize,nameinlink]{cleveref}
\usepackage{tikz}
\usepackage{tikz-dependency}
\usepackage{adjustbox}
\usepackage{todonotes}
\usepackage[multiple]{footmisc}
\newcommand{\bertb}{BERT\textsubscript{base}}

\newcommand{\xlmr}{\texttt{XLM-R}\textsubscript{large}}
\newcommand{\escolmr}{\texttt{ESCOXLM-R}}

\newcommand{\std}[2]{{#1}{\footnotesize$\pm${#2}}}

\newcommand{\lr}[2]{{#1}{$e^{-\text{#2}}$}}

\usepackage[T1]{fontenc}

\usepackage[utf8]{inputenc}

\usepackage{microtype}

\usepackage{inconsolata}

\title{ESCOXLM-R: Multilingual Taxonomy-driven Pre-training\\ for the Job Market Domain}

 \author{Mike Zhang\textsuperscript{\faCompass} \and
         Rob van der Goot\textsuperscript{\faCompass} \and
         Barbara Plank\textsuperscript{\faCompass}\textsuperscript{\faMountain} \\
  \textsuperscript{\faCompass}Department of Computer Science, IT University of Copenhagen, Denmark\\
  \textsuperscript{\faMountain}MaiNLP, Center for Information and Language Processing, LMU Munich, Germany \\
{\tt \{mikz, robv\}@itu.dk } \hspace{2em} {\tt b.plank@lmu.de}}

\begin{document}
\maketitle
\begin{abstract}
\looseness=-1
The increasing number of benchmarks for Natural Language Processing (NLP) tasks in the computational job market domain highlights the demand for methods that can handle job-related tasks such as skill extraction, skill classification, job title classification, and de-identification. While some approaches have been developed that are specific to the job market domain, there is a lack of generalized, multilingual models and benchmarks for these tasks. In this study, we introduce a language model called \escolmr{}, based on \xlmr{}, which uses domain-adaptive pre-training on the European Skills, Competences, Qualifications and Occupations (ESCO) taxonomy, covering 27 languages. The pre-training objectives for \escolmr{} include dynamic masked language modeling and a novel additional objective for inducing multilingual taxonomical ESCO relations.
We comprehensively evaluate the performance of \escolmr{} on 6 sequence labeling and 3 classification tasks in 4 languages and find that it achieves state-of-the-art results on 6 out of 9 datasets. Our analysis reveals that \escolmr{} performs better on short spans and outperforms \xlmr{} on entity-level and surface-level span-F1, likely due to ESCO containing short skill and occupation titles, and encoding information on the entity-level.
\end{abstract}


\section{Introduction}
\looseness=-1
The dynamic nature of labor markets, driven by technological changes, migration, and digitization, has resulted in a significant amount of job advertisement data (JAD) being made available on various platforms to attract qualified candidates~\cite{brynjolfsson2011race,brynjolfsson2014second,balog2012expertise}. This has led to an increase in tasks related to JAD, including skill extraction~\cite{kivimaki-etal-2013-graph,zhao2015skill,sayfullina2018learning,smith2019syntax,tamburri2020dataops,shi2020salience, chernova2020occupational,bhola-etal-2020-retrieving,zhang-etal-2022-skillspan,zhang-jensen-plank:2022:LREC, zhang2022skill,green-maynard-lin:2022:LREC,gnehm-bhlmann-clematide:2022:LREC,beauchemin2022fijo,decorte2022design, goyal-etal-2023-jobxmlc}, skill classification~\cite{decorte2022design,zhang-jensen-plank:2022:LREC}, job title classification~\cite{javed2015carotene,javed2016towards,decorte2021jobbert,green-maynard-lin:2022:LREC}, de-identification of entities in job postings~\cite{jensen2021identification}, and multilingual skill entity linking~\cite{esco-2022}.

\looseness=-1
While some previous studies have focused on JAD in non-English languages~\cite{zhang-jensen-plank:2022:LREC,gnehm-bhlmann-clematide:2022:LREC,beauchemin2022fijo}, their baselines have typically relied on language-specific models, either using domain-adaptive pre-training (DAPT; \citealp{gururangan2020don}) or off-the-shelf models. The lack of comprehensive, open-source JAD data in various languages makes it difficult to fully pre-train a language model (LM) using such data. In this work, we seek external resources that can help improve the multilingual performance on the JAD domain. We use the ESCO taxonomy~\cite{le2014esco}, which is a standardized system for describing and categorizing the skills, competences, qualifications, and occupations of workers in the European Union. The ESCO taxonomy, which has been curated by humans, covers over 13,000 skills and 3,000 occupations in 27 languages. Therefore, we seek to answer: \emph{To what extent can we leverage the ESCO taxonomy to pre-train a domain-specific and language-agnostic model for the computational job market domain?}

\begin{figure*}
    \centering
    \includegraphics[width=\linewidth]{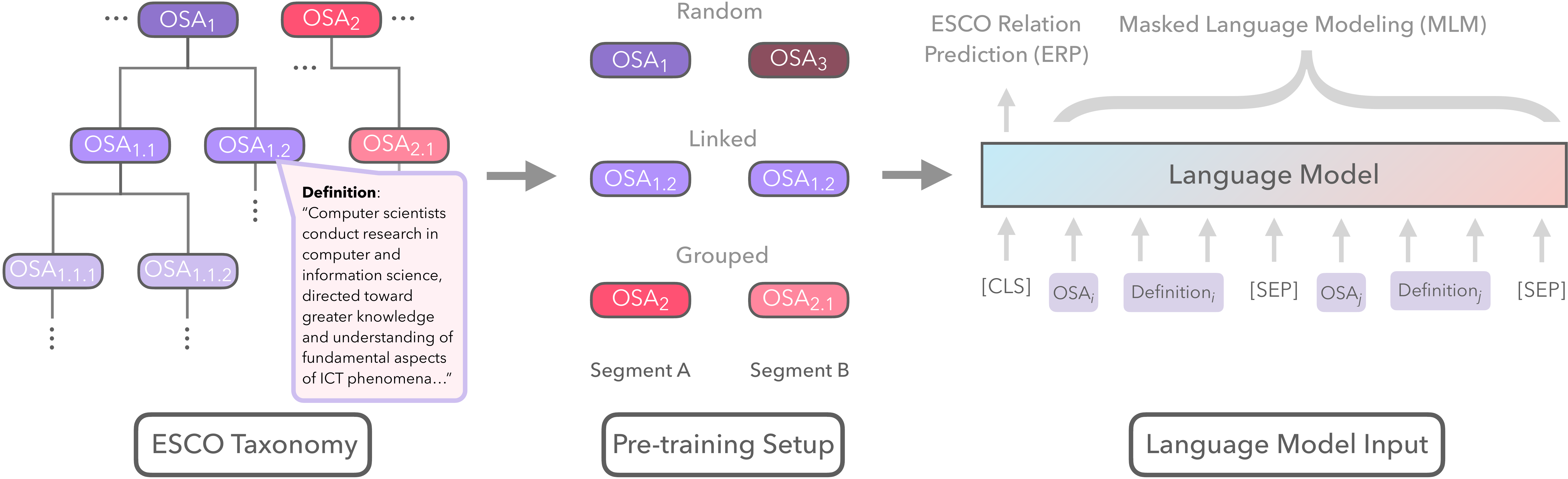}
    \caption{\textbf{ESCO Pre-training Objective}: From left to right, the figure illustrates the hierarchical structure of the ESCO taxonomy, which consists of occupations, skills, and aliases (OSA). Each OSA includes a definition. For the purposes of this study, we consider aliases of occupations to have the same definition as the occupation itself. In the middle of the figure, we show our pre-training setup. Pre-training instances are uniformly sampled in three ways: randomly, linked, or grouped (this is defined in~\cref{subsec:pretraining}). The selected instances (can be in different languages) are then fed to the language model, along with its description. We have two pre-training objectives: the regular MLM objective, and a new ESCO relation prediction objective, in which the goal is to predict which group the sampled instances belong to (Random, Linked, or Grouped).}
    \label{fig:1}
\end{figure*}

\looseness=-1
In this work, we release the first multilingual JAD-related model named \escolmr{}, a language model based on \xlmr{} that incorporates data from the ESCO taxonomy through the use of two pre-training objectives (\cref{fig:1}): Masked Language Modeling (MLM) and a novel ESCO relation prediction task (\cref{sec:escolmr}). We evaluate \escolmr{} on 9 JAD-related datasets in 4 different languages covering 2 NLP tasks (\cref{sec:exp}). Our results show that \escolmr{} outperforms previous state-of-the-art (SOTA) on 6 out of 9 datasets (\cref{sec:results}). In addition, our fine-grained analysis reveals that \escolmr{} performs better on short spans compared to \xlmr{}, and consistently outperforms \xlmr{} on entity-level and surface-level span-F1 (\cref{sec:disc}).

\paragraph{Contributions} 
\looseness=-1
In this work, we present and release the following:
\begin{itemize}
   \itemsep0em
    \item \escolmr{}, an \xlmr-based model, which utilizes domain-adaptive pre-training on the 27 languages from ESCO.\footnote{The code for \escolmr{} is available as open-source: \url{https://github.com/mainlp/escoxlmr}. We further release \escolmr{} under an Apache License 2.0 on HuggingFace: \url{https://huggingface.co/jjzha/esco-xlm-roberta-large}.}
    \item The largest JAD evaluation study to date on 3 job-related tasks, comprising 9 datasets in 4 languages and 4 models.
    \item A fine-grained analysis of \escolmr's performance on different span lengths, and emerging entities (i.e., recognition of entities in the long tail).
\end{itemize}

\section{ESCOXLM-R}\label{sec:escolmr}

\paragraph{Preliminaries}

\looseness=-1
In the context of pre-training, an LM is trained using a large number of unlabeled documents, $\mathcal{X} = {X^{(i)}}$, and consists of two main functions: $f_\text{encoder}(.)$, which maps a sequence of tokens $X = (x_1, x_2, ..., x_t)$ to a contextualized vector representation for each token, represented as $(h_1, h_2, ..., h_t)$, and $f_\text{head}(.)$, the output layer that takes these representations and performs a specific task, such as pre-training in a self-supervised manner or fine-tuning on a downstream application. For example, BERT~\cite{devlin2019bert} is pre-trained using two objectives: MLM and Next Sentence Prediction (NSP). In MLM, a portion of tokens in a sequence $X$ is masked and the model must predict the original tokens from the masked input. In the NSP objective, the model takes in two segments $(X_\text{A}, X_\text{B})$ and predicts whether segment $X_\text{B}$ follows $X_\text{A}$. RoBERTa~\cite{liu2019roberta} is a variation of BERT that uses dynamic MLM, in which the masking pattern is generated each time a sequence is fed to the LM, and does not use the NSP task.

\begin{figure*}[ht]
    \centering
    \includegraphics[width=\linewidth]{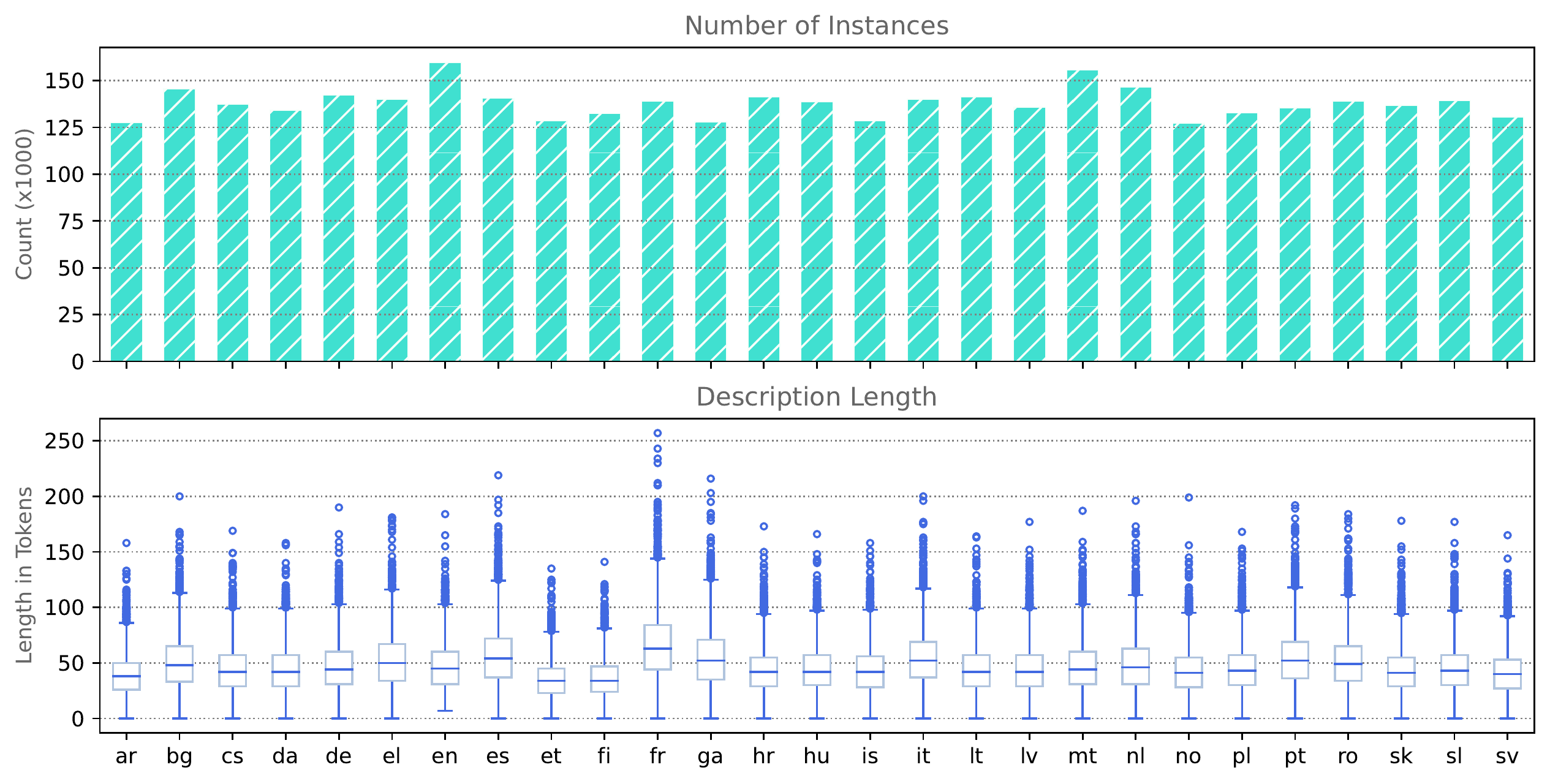}
    \caption{\textbf{Statistics of Pre-training Data.} The ESCO dataset contains descriptions in 27 languages, with a combined total of approximately 3.72 million descriptions (i.e., instances). On average, there are around 130,000 descriptions per language. The average length of each description is 26.3 tokens, with some descriptions reaching a maximum length of 150 or more tokens, as shown by the outliers in the boxplot.}
    \label{fig:esco_stats}
\end{figure*}

\paragraph{Multilinguality} Both BERT and RoBERTa have been extended to support multiple languages, resulting in multilingual BERT (mBERT; \citealp{devlin2019bert}) and XLM-RoBERTa (\texttt{XLM-R}; \citealp{conneau2020unsupervised}). XLM-R was found to outperform mBERT on many tasks (e.g.,~\citealp{conneau2020unsupervised,hu2020xtreme, lauscher2020zero}) due to careful tuning, sampling, and scaling to larger amounts of textual data. Because of this, our \escolmr{} model is based on \xlmr{}.

\subsection{European Skills, Competences, Qualifications and Occupations Taxonomy}\label{subsec:esco}
\looseness=-1
The European Skills, Competences, Qualifications, and Occupations (ESCO;~\citealp{le2014esco}) taxonomy is a standardized system for describing and categorizing the skills, competences, qualifications, and occupations of workers in the European Union (EU). It is designed to serve as a common language for the description of skills and qualifications across the EU, facilitating the mobility of workers by providing a common reference point for the recognition of qualifications and occupations. The taxonomy is developed and maintained by the European Commission and is based on the International Classification of Occupations and the International Standard Classification of Education. It includes 27 European languages: Bulgarian (ar), Czech (cs), Danish (da), German (de), Greek (el), English (en), Spanish (es), Estonian (et), Finnish (fi), French (fr), Gaelic (ga), Croatian (hr), Hungarian (hu), Icelandic (is), Italian (it), Lithuanian (lt), Latvian (lv), Maltese (mt), Dutch (nl), Norwegian (no), Polish (pl), Portuguese (pt), Romanian (ro), Slovak (sk), Slovenian (sl), Swedish (sv), and Arabic (ar). Currently, it describes 3,008 occupations and 13,890 skills/competences (SKC) in all 27 languages.\footnote{Note that ESCO now also includes Ukrainian, but this model was trained before that inclusion. We use the ESCO V1.0.9 API to extract the data. ESCO contains an Apache 2.0 and a European Union Public License 1.2.}

The ESCO taxonomy includes a hierarchical structure with links between occupations, skills, and aliases (OSA). In this work, we focus on the occupation pages and extract the following information from the taxonomy:\footnote{An example of the extracted information can be found in~\cref{json-example} (\cref{sec:appendixa}), and the original page can be accessed at \url{https://bit.ly/3DY1zsX}.}
\begin{itemize}
\itemsep0em
\item \texttt{ESCO Code}: The taxonomy code for the specific occupation or SKC.
\item \texttt{Occupation Label}: The preferred occupation name (i.e., title of the occupation).
\item \texttt{Occupation Description/Definition}: A description of the responsibilities of the specific occupation.
\item \texttt{Major Group Name}: The name of the overarching group to which the occupation belongs, e.g., ``Veterinarians'' for the occupation ``animal therapist''.
\item \texttt{Alternative Labels}: Aliases for the specific occupation, e.g., ``animal rehab therapist'' for the occupation ``animal therapist''.
\item \texttt{Essential Skills}: All necessary SKCs for the occupation, including descriptions of these.
\item \texttt{Optional Skills}: All optional SKCs for the occupation, including descriptions of these.
\end{itemize}

In~\cref{fig:esco_stats}, we present the distribution of pre-training instances and the mean description lengths for each language in the ESCO taxonomy. Note that the number of descriptions is not the same for all languages, and we do not count empty descriptions (i.e., missing translations) for certain occupations or SKCs.

\begin{figure}[t]
    \centering
    \includegraphics[width=\linewidth]{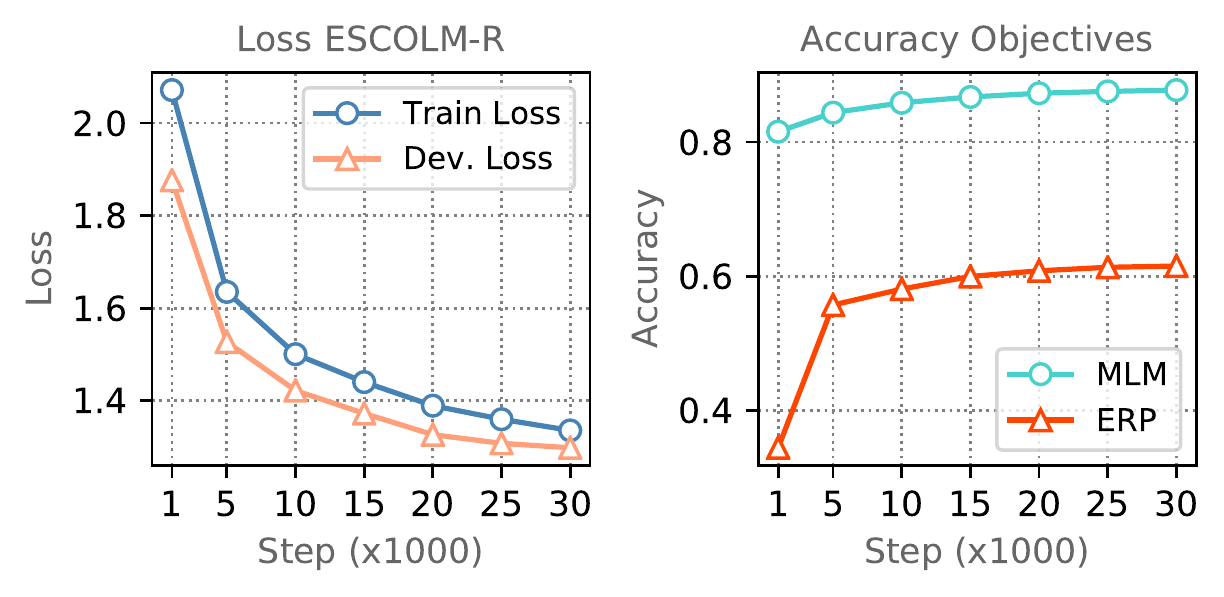}
    \caption{\textbf{Pre-Training Statistics.} The final log loss for the training set is 1.34, while the log loss for the development set is 1.30. The MLM accuracy is 84.3\%, while the Entity Relationship Prediction (ERP) accuracy is 60.0\%. These results were obtained after approximately 1.04 epochs of training on the total data. }
    \label{fig:pretraining}
\end{figure}

\subsection{Pre-training Setup}\label{subsec:pretraining}
\looseness=-1
To improve our \xlmr-based model, we employ domain-adaptive pre-training techniques as described in previous work such as~\citet{alsentzer2019publicly,han-eisenstein-2019-unsupervised,leebiobert,gururangan2020don,nguyen2020bertweet}. Given the limited amount of training data (3.72M sentences), we utilize the \xlmr{} checkpoint provided by the HuggingFace library~\cite{wolf-etal-2020-transformers} as a starting point.\footnote{\url{https://huggingface.co/xlm-roberta-large}} Our aim is to fine-tune the model to internalize domain-specific knowledge related to occupation and SKCs, while maintaining its general knowledge acquired during the original pre-training phase.

\looseness=-1
We introduce a novel self-supervised pre-training objective for \escolmr{}, inspired by LinkBERT from~\citet{yasunaga-etal-2022-linkbert}. We view the ESCO taxonomy as a graph of occupations and SKCs (\cref{fig:1}), with links between occupations or occupations and SKCs in various languages. By placing similar occupations or SKCs in the same context window and in different languages, we can learn from the links between (occupation $\leftrightarrow$ occupation) and (occupation $\leftrightarrow$ SKCs) in different languages for true cross-lingual pre-training. In addition to the MLM pre-training objective, which is used to learn concepts within contexts, we introduce another objective called ESCO Relation Prediction (ERP) to internalize knowledge of connections within the taxonomy in the LM. We take an anchor concept ($C_\text{A}$) by concatenating it with its description ($X_\text{A}$) from the ESCO taxonomy and sample an additional concept ($C_\text{B}$) concatenated with its description ($X_\text{B}$) to create LM input \texttt{[CLS]} $C_\text{A} X_\text{A}$ \texttt{[SEP]} $C_\text{B} X_\text{B}$ \texttt{[SEP]}.\footnote{The special tokens used in this example follow the naming convention of BERT for readability, \texttt{[CLS]} and \texttt{[SEP]}. However, since we use \xlmr{} there are different special tokens: \texttt{<s>} as the beginning of the sequence, \texttt{</s>} as the \texttt{SEP} token, and \texttt{</s></s>} as segment separators. Formally, given the example in the text: \texttt{<s> $C_\text{A}X_\text{A}$ </s></s> $C_\text{B}X_\text{B}$ </s>}.} We sample $C_\text{B} X_\text{B}$ in three ways with uniform probability:

\begin{enumerate}
    \itemsep0em
    \item \emph{Random}: We randomly sample $C_\text{B}X_\text{B}$ from the ESCO taxonomy, in any language;
    \item \emph{Linked}: We sample $C_\text{B}X_\text{B}$ in any language from the same occupation page, for example, an ``animal therapist'' (or an alias of the ``animal therapist'', e.g., ``animal rehab therapist'') should have knowledge of ``animal behavior'';
    \item \emph{Grouped}: We sample $C_\text{B}X_\text{B}$ from the same major group in any language. For the same example ``animal therapist'', it comes from major group 2: Professionals $\rightarrow$ group 22: Health professionals. Several other concepts, e.g., ``Nursing professionals'' fall under this major group.
\end{enumerate}

\begin{table*}[ht]
\centering
\setlength{\arrayrulewidth}{0.25mm}
\resizebox{\linewidth}{!}{
\begin{tabular}[t]{|l|l|l|l|l|l|l|rrr|}
\hline


Dataset Name           & Lang.\          & Loc.\   & License   & Task           & Metric        & Input Type    &   Train                   &      Dev.             &   Test \\
\hline\hline
\textsc{SkillSpan}     & en                & *       & CC-BY-4.0    & SL             & Span-F1       & Sentences       &   5,866               &      3,992                    &   4,680          \\
\textsc{Sayfullina}    & en                & UK      & Unknown    & SL             & Span-F1       & Sentences       &   3,706               &      1,854                    &   1,853          \\
\textsc{Green}         & en                & UK      & CC-BY-4.0    & SL             & Span-F1       & Sentences       &   8,670               &      963                      &   336            \\
\textsc{JobStack}      & en                & *       & RLT          & SL             & Span-F1       & Sentences       &   18,055              &      2,082                    &   2,092          \\
\textsc{Bhola}         & en                & SG      & CC-BY-4.0          & MLC            & MRR           & Documents       &   16,238              &      2,030                       &   2,030          \\
\textsc{Kompetencer}   & en                & DK      & CC-BY-4.0    & MCC            & W.\ Macro-F1  & Skills          &   9,472                &      1,577                    &   1,578         \\
\textsc{Kompetencer}   & \textbf{da}       & DK      & CC-BY-4.0    & MCC            & W.\ Macro-F1  & Skills          &   138                 &      -                    &   784    \\
\textsc{Gnehm}         & \textbf{de}       & CH      & CC-BY-NC-SA-4.0     & SL             & Span-F1       & Sentences        &   22,134              &      2,679                     &   2,943          \\
\textsc{Fijo}          & \textbf{fr}       & FR      & Unknown    & SL             & Span-F1       & Sentences       &   399                 &      50                       &   50             \\

\hline
\end{tabular}}
\caption{\textbf{Dataset Statistics.} We show statistics for all 9 JAD datasets. There are 6 datasets in English and 3 in other languages (Danish, German, and French). We indicate the location the JAD originates from (whenever applicable, * indicates it comes from a variety of countries). We indicate the license of the dataset. Most of the task types consist of sequence labeling (e.g., span extraction, Named Entity Recognition, soft skill tagging). To maintain consistency, we use a single metric for each task type: Sequence Labeling (SL), Multilabel Classification (MLC), and Multiclass Classification (MCC). For \textsc{Kompetencer}, the statistics are provided in brackets for the Danish language.
}
\label{tab:num_post}
\end{table*}

\paragraph{Pre-training\ Objectives} 
The LM is trained using two objectives. First is the MLM objective, and the second is the ERP objective, where the task is to classify the relation $r$ of the \texttt{[CLS]} token in \texttt{[CLS]} $C_\text{A} X_\text{A}$ \texttt{[SEP]} $C_\text{B} X_\text{B}$ \texttt{[SEP]} ($r \in {\text{Random}, \text{Linked}, \text{Grouped}}$). The rationale behind this is to encourage the model to learn the relevance between concepts in the ESCO taxonomy. We formalize the objectives in~\cref{eq:pretrain}:

\begin{equation}\label{eq:pretrain}
\begin{aligned}
\mathcal{L} &= \mathcal{L}_{\mathrm{MLM}} + \mathcal{L}_{\mathrm{ERP}} \\
&=-\sum_{i} \log p\left(x_{i} \mid \mathbf{h}_{i}\right) - \log  p\left(r \mid \mathbf{h}_{\text{\texttt{[CLS]}}}\right),
\end{aligned}
\end{equation}

\looseness=-1
we define the overall loss $\mathcal{L}$ as the sum of the MLM loss $\mathcal{L}_{\mathrm{MLM}}$ and the ERP loss $\mathcal{L}_{\mathrm{ERP}}$. The MLM loss is calculated as the negative log probability of the input token $x_i$ given the representation $\mathbf{h}_i$. Similarly, the ERP loss is the negative log probability of the relationship $r$ given the representation of the start-token $\mathbf{h}_{\text{\texttt{[CLS]}}}$. In our implementation, we use \xlmr{} and classify the start-token \texttt{[CLS]} for ERP to improve the model's ability to capture the relationships between ESCO occupations and skills.

\paragraph{Implementation}
\looseness=-1
For optimization we follow~\cite{yasunaga-etal-2022-linkbert}, we use the AdamW~\cite{loshchilov2017decoupled} optimizer with ($\beta_1$, $\beta_2$) = (0.9, 0.98). We warm up the learning rate \lr{1}{5} for a ratio of 6\% and then linearly decay it. The model is trained for 30K steps, which is equivalent to one epoch over the data, and the training process takes 33 hours on one A100 GPU with tf32. We use a development set comprising 1\% of the data for evaluation. In~\cref{fig:pretraining}, the pre-training loss and performance on the dev.\ set are plotted, it can be seen that the accuracy plateaus at 30K steps. Though the train and development loss hint that further gains could be obtained on the pretraining objective, we found through empirical analysis on downstream tasks that 30K steps performs best.

\section{Experimental Setup}\label{sec:exp}
\cref{tab:num_post} provides the details of the downstream datasets used in this study. Most of the datasets are in EN, with a smaller number in DA, DE, and FR. For each dataset, a brief description and the corresponding best-performing models are given. We put examples of each dataset (apart from JobStack due to the license) in~\cref{data:examples}.

\paragraph{\textsc{SkillSpan}~\cite{zhang-etal-2022-skillspan}}
\looseness=-1
The job posting dataset includes annotations for skills and knowledge, derived from the ESCO taxonomy. The best model in the relevant paper, JobBERT, was retrained using a DAPT approach on a dataset of 3.2 million EN job posting sentences. This is the best-performing model which we will compare against.

\paragraph{\textsc{Kompetencer}~\cite{zhang-jensen-plank:2022:LREC}}
\looseness=-1
This dataset is used to evaluate models on the task of classifying skills according to their ESCO taxonomy code. It includes EN and DA splits, with the EN set derived from \textsc{SkillSpan}. There are three experimental setups for evaluation: fully supervised with EN data, zero-shot classification (EN$\rightarrow$DA), and few-shot classification (a few DA instances). The best-performing model in this work is RemBERT~\cite{chung2020rethinking}, which obtains the highest weighted macro-F1 for both EN and DA. In this work, we use setup 1 and 3, where all available data is used.

\begin{table*}[ht]
\centering
\setlength{\arrayrulewidth}{0.25mm}
\resizebox{\linewidth}{!}{
\begin{tabular}[t]{|l|ll|llr|ll|}
\hline

 Dataset                      &     Lang.       & Metric         &           Prev. SOTA                                 &       \xlmr                 &          \xlmr{} (+ DAPT)              & \escolmr{}                & \multicolumn{1}{c|}{$\Delta$} \\
 \hline\hline
\textsc{SkillSpan}            &     EN             & Span-F1        &           \std{58.9}{4.5}                               &       \std{59.7}{4.6}     &         \std{62.0}{4.0}                 & \textbf{\std{62.6}{3.7}}     & $+$3.7          \\
\textsc{Sayfullina}           &     EN             & Span-F1        &           \std{73.1}{2.1}                               &       \std{89.9}{0.5}       &          \std{90.6}{0.4}              & \textbf{\std{92.2}{0.2}}     & $+$19.1         \\
\textsc{Green}                &     EN             & Span-F1        &           \std{31.8}{*}                                 &       \std{49.0}{2.4}       &          \std{47.5}{0.7}              & \textbf{\std{51.2}{2.1}}     & $+$19.4         \\
\textsc{Jobstack}             &     EN             & Span-F1        &           \textbf{\std{82.1}{0.8}}                      &       \std{81.2}{0.6}       &          \std{80.4}{0.7}              & \std{82.0}{0.7}              & $-$0.1          \\
\textsc{Kompetencer}          &     EN             & W. Macro-F1    &           \std{62.8}{2.8}                               &       \std{59.0}{9.5}       &         \textbf{\std{64.3}{0.5}}      & \std{63.5}{1.3}              & $-$0.7          \\
\textsc{Bhola}                &     EN             & MRR            &           \std{90.2}{0.2}                               &       \std{90.5}{0.3}       &          \std{90.0}{0.3}              & \textbf{\std{90.7}{0.2}}     & $+$0.5          \\
 \hline\hline
\textsc{Gnehm}                &     DE            & Span-F1        &           \std{86.7}{0.4}                               &       \std{87.1}{0.4}       &          \std{86.8}{0.2}            & \textbf{\std{88.4}{0.5}}     & $+$1.7          \\
\textsc{Fijo}                 &     FR             & Span-F1        &           \std{31.7}{2.3}                               &       \std{41.8}{2.0}       &         \std{41.7}{0.7}             & \textbf{\std{42.0}{2.3}}     & $+$10.3         \\
\textsc{Kompetencer}          &     DA             & W. Macro-F1    &           \std{45.3}{1.5}                               &       \std{41.2}{9.8}       &         \textbf{\std{45.6}{0.8}}    & \std{45.0}{1.4}              & $-$0.3          \\
\hline
\end{tabular}}
\caption{\textbf{Results of Experiments.} The datasets and models are described in~\cref{sec:exp}. We re-train the best-performing models of all papers to give us the standard deviation. The best-performing model is in bold. The difference in performance between \escolmr{} and the previous SOTA is shown as $\Delta$.
Note (*) that the results for \textsc{Green} are based on a CRF model where the data has been pre-split, and therefore, there is no standard deviation.}
\label{tab:res}
\end{table*}

\paragraph{\textsc{Bhola}~\cite{bhola-etal-2020-retrieving}}
The task of this EN job posting dataset is multilabel classification: Predicting a list of necessary skills in for a given job description. It was collected from a Singaporean government website. It includes job requirements and responsibilities as data fields. Pre-processing steps included lowercasing, stopword removal, and rare word removal. Their model is BERT with a bottleneck layer~\cite{liu2017deep}. In our work, the bottleneck layer is not used and no additional training data is generated through bootstrapping. To keep comparison fair, we re-train their model without the additional layer and bootstrapping. We use Mean Reciprocal Rank as the main results metric.

\paragraph{\textsc{Sayfullina}~\cite{sayfullina2018learning}}
\looseness=-1
This dataset is used for soft skill prediction, a sequence labeling problem. Soft skills are personal qualities that contribute to success, such as ``team working'', ``being dynamic'', and ``independent''. The models for this dataset include a CNN~\cite{kim-2014-convolutional}, an LSTM~\cite{hochreiter1997long}, and a Hierarchical Attention Network~\cite{yang2016hierarchical}. We compare to their best-performing LSTM model.

\paragraph{\textsc{Green}~\cite{green-maynard-lin:2022:LREC}}
\looseness=-1
A sentence-level sequence labeling task involving labeling skills, qualifications, job domain, experience, and occupation labels. The job positions in the dataset are from the United Kingdom. The industries represented in the data vary and include IT, finance, healthcare, and sales. Their model for this task is a Conditional Random Field~\cite{lafferty2001conditional} model.

\paragraph{\textsc{JobStack}~\cite{jensen2021identification}}
\looseness=-1
This corpus is used for de-identifying personal data in job vacancies on Stack Overflow. The task involves sequence labeling and predicting Organization, Location, Name, Profession, and Contact details labels. The best-performing model for this task is a transformer-based~\cite{vaswani2017attention} model trained in a multi-task learning setting. \citet{jensen2021identification} propose to use the I2B2/UTHealth corpus, which is a medical de-identification task~\cite{stubbs2015annotating}, as auxiliary data, which showed improvement over their baselines.

\paragraph{\textsc{Gnehm}~\cite{gnehm-bhlmann-clematide:2022:LREC}}
\looseness=-1
A Swiss-German job ad dataset where the task is Information and Communications Technology (ICT)-related entity recognition, these could be ICT tasks, technology stack, responsibilities, and so forth. The used dataset is a combination of two other Swiss datasets namely the Swiss Job Market Monitor and an online job ad dataset~\cite{gnehm2020text, buchmann2022swiss}. Their model is dubbed JobGBERT and is based on DAPT with German \bertb{}~\cite{chan2020german}.

\paragraph{\textsc{Fijo}~\cite{beauchemin2022fijo}}
\looseness=-1
A French job ad dataset with the task of labeling skill types using a sequence labeling approach. The skill groups are based on the AQESSS public skills repositories and proprietary skill sets provided by their collaborators. These skill types are divided into four categories: ``Thoughts'', ``Results'', ``Relational'', and ``Personal''. The best-performing model for this task is CamemBERT~\cite{martin2020camembert}.

\begin{figure*}[t]
    \centering
    \includegraphics[width=\linewidth]{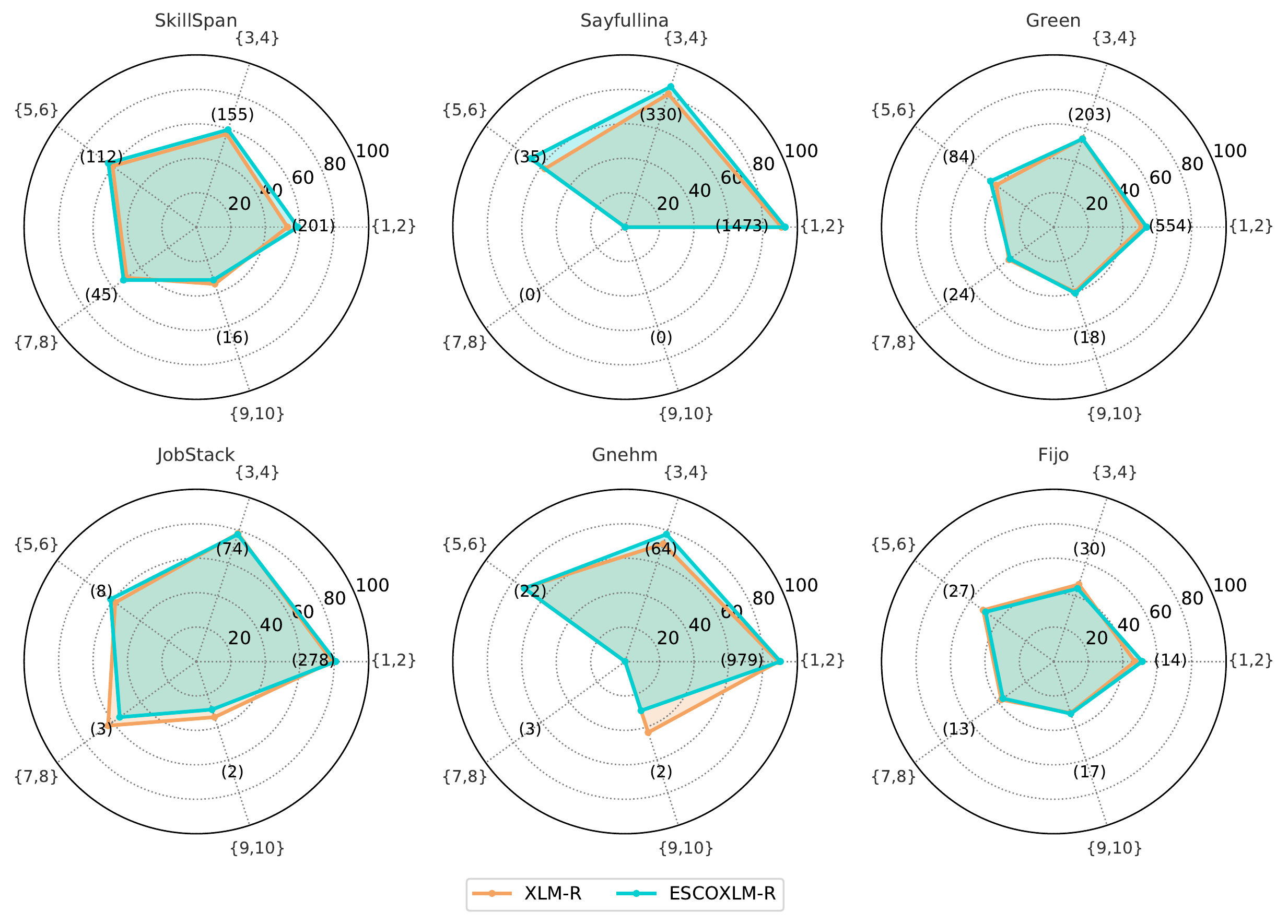}
    \caption{\textbf{Radar Charts of Span-F1 performance by Span Token Length.} We show the performance of \xlmr{} and \escolmr{} on different span lengths, we bucketed the performances of both models according to the length of the spans, up to 10 tokens, and presented the average performance over five random seeds. We did not include error bars in these plots. Note that in some plots, there are no instances in certain buckets (e.g., \textsc{Sayfullina} with 7-8, 9-10). Also, some outer rings only go up to 60 span F1, rather than 100.}
    \label{fig:radar}
\end{figure*}

\begin{table*}
    \centering
    \begin{tabular}{|l|l|ll|ll|}
    \hline
    Dataset              &   Ratio         & \multicolumn{2}{c|}{Span-F1 (Entity)}                  &     \multicolumn{2}{c|}{Span-F1 (Surface)}        \\\hline\hline
                         &                       &    \texttt{XLM-R}             & \escolmr{}                    &  \texttt{XLM-R}                  & \escolmr{}            \\\hline\hline
    \textsc{SkillSpan}   &   0.90               &  \std{59.9}{7.9}     &  \textbf{\std{61.6}{6.6}}            &  \std{56.4}{5.7}        & \textbf{\std{57.9}{4.3}}     \\
    \textsc{Sayfullina}  &   0.22                &  \std{94.0}{0.2}     &  \textbf{\std{95.7}{0.3}}            &  \std{82.8}{0.6}        & \textbf{\std{87.2}{0.7}}     \\
    \textsc{Green}       &   0.79                &  \std{50.3}{2.4}     &  \textbf{\std{53.1}{2.1}}            &  \std{49.2}{2.4}        &  \textbf{\std{52.0}{2.1}}    \\
    \textsc{Jobstack}    &   0.41                &  \std{85.6}{0.7}     &  \textbf{\std{86.4}{0.5}}            &  \std{78.4}{1.2}        & \textbf{\std{79.8}{0.7}}     \\
    \textsc{Gnehm}       &   0.53                &  \std{89.3}{0.3}     &  \textbf{\std{89.6}{0.4}}            &  \std{87.3}{0.3}        & \textbf{\std{87.8}{0.6}}     \\
    \textsc{Fijo}        &   0.77                &  \std{34.4}{2.9}     &  \textbf{\std{35.7}{1.1}}            &  \std{34.4}{1.1}        &  \textbf{\std{35.7}{1.1}}     \\\hline
    \end{tabular}
    \caption{\textbf{Entity vs.\ Surface-level span-F1 on Test.} In this table, the performance of two systems, \xlmr{} and \escolmr{}, was measured using entity-level and surface-level span-F1 scores. Entity-level span-F1 measures precision, recall, and harmonic mean at the entity level, while surface-level span-F1 measures a system's ability to recognize a range of entities. We include the ratio of surface entities to total entities in each \emph{training} set, with a higher ratio indicating more variety (a ratio of 1.00 indicates all entities are unique). 
    }
    \label{tab:entsurf}
\end{table*}

\section{Results}\label{sec:results}
\looseness=-1
The results of the models are presented in~\cref{tab:res}.  To evaluate the performance, four different models are used in total: \escolmr{}, the best-performing model originally reported in the relevant paper for the downstream task, vanilla \xlmr{}, and an \xlmr{} model that we continuously pre-trained using only MLM (DAPT; excluding the ERP objective) using the same pre-training hyperparameters as \escolmr{}. For more information regarding the hyperparameters of fine-tuning, we refer to~\cref{app:finetuning} (\cref{tab:hyperparameters}).


\looseness=-1
\paragraph{English} \escolmr{} is the best-performing model in 4 out of 6 EN datasets. The largest improvement compared to the previous SOTA is observed in \textsc{Sayfullina} and \textsc{Green}, with over 19 F1 points. In 3 out of 4 datasets, \escolmr{} has the overall lower standard deviation. For \textsc{Jobstack}, the previous SOTA performs best, and for \textsc{Kompetencer}, \xlmr{} (+ DAPT) has the highest performance.

\looseness=-1
\paragraph{Non-English} In 2 out of 3 datasets, \escolmr{} improves over the previous SOTA, with the largest absolute difference on French \textsc{Fijo} with 10.3 F1 points. In the Danish subset of \textsc{Kompetencer}, \xlmr{} (+ DAPT) has higher performance than \escolmr{}. Next, we will discuss potential reasons for these differences. 

\looseness=-1
\subsection{Analysis} We highlight that the performance gains of \escolmr{} are generally much larger than any of the losses, indicating a largely positive effect of training on ESCO. The improved performance of \escolmr{} on JAD datasets in~\cref{tab:res} is likely due to the focus on tasks with token-level annotation (i.e., sequence labeling). This suggests that pre-training on the ESCO taxonomy is particularly useful for these types of tasks. The under-performance of \escolmr{} on the \textsc{Kompetencer} dataset in both EN and DA may be because the task involves predicting the ESCO taxonomy code for a given skill \emph{without context}, where we expect ESCO to particularly help with tasks where having context is relevant. We suspect applying DAPT and ERP on ESCO specifically improves recognizing entities that are uncommon.  On the other hand, the poor performance on the \textsc{Jobstack} dataset may be due to the task of predicting various named entities, such as organizations and locations. By manual inspection, we found that ESCO does not contain entities related to organizations, locations, or persons, thus this reveals that there is a lack of relevant pre-training information to \textsc{Jobstack}. 

\section{Discussion}\label{sec:disc}

\subsection{Performance on Span Length}
We seek to determine whether the difference in performance between the \escolmr{} and \xlmr{} models is due to shorter spans, and to what extent. One application of predicting short spans well is the rise of technologies, for which the names are usually short in length.
~\citet{zhang2022skill} observes that skills described in the ESCO dataset are typically short, with a median length of approximately 3 tokens. We compare the average performance of both models on the test sets of each dataset, where span-F1 is used as measurement. We group gold spans into buckets of lengths 1-2, 3-4, 5-6, 7-8, and 9-10, and present the span-F1 for each model (\xlmr{} vs. \escolmr{}) in each bucket.

Shown in~\cref{fig:radar}, \escolmr{} outperforms \xlmr{} on shorter spans (i.e., 1-2 or 3-4) in 6 out of the 6 datasets, suggesting that pre-training on ESCO is beneficial for predicting short spans. However, there is a slight decline in performance on some datasets (e.g., \textsc{Skillspan}, \textsc{Jobstack}, and \textsc{Gnehm}) when the spans are longer (i.e., 7-8 or 9-10). It is worth noting that the number of instances in these longer span buckets is lower, and therefore errors may be less apparent in terms of their impact on overall performance.

\subsection{Entity-F1 vs. Surface-F1}

In this analysis, we adopt the evaluation method used in the W-NUT shared task on Novel and Emerging Entity Recognition~\cite{derczynski-etal-2017-results}. In this shared task, systems are evaluated using two measures: entity span-F1 and surface span-F1. Entity span-F1 assesses the precision, recall, and harmonic mean (F1) of the systems at the entity level, while surface span-F1 assesses their ability to correctly recognize a diverse range of entities, rather than just the most frequent surface forms. 
This means surface span-F1 counts entity types, in contrast to entity tokens in the standard entity span-F1 metric. 

As shown in~\cref{tab:entsurf}, we first calculate the ratio of unique entities and total entities in each relevant train set (i.e., datasets where we do span labeling). A higher ratio number indicates a wider variety of spans. Both \xlmr{} and \escolmr{} tend to have lower performance when variety gets high (above 0.75). In addition, there are 2 datasets (\textsc{Sayfullina}, \textsc{Jobstack}) where we see a low variety of spans and large discrepancy between performance of entity span-F1 and surface span-F1. This difference is lower for \escolmr{} (especially in \textsc{Sayfullina}) suggesting that pre-training on ESCO helps predicting uncommon entities.


It is also noteworthy that the standard deviations for the scores at the entity span-F1 are generally lower than those for the surface span-F1. This suggests that the results for the entity span-F1 scores are more consistent across different runs, likely due to recognizing common entities more.

Overall, \escolmr{} consistently outperforms \xlmr{} in both the entity-level and surface-level F1 scores, indicating the benefits of using the ESCO dataset for pre-training on JAD tasks.

\section{Related Work}
\looseness=-1
To the best of our knowledge, we are the first to internalize an LM with ESCO for job-related NLP tasks. There are, however, several works that integrate factual knowledge (i.e., knowledge graphs/bases) into an LM. 
\citet{peters-etal-2019-knowledge} integrates multiple knowledge bases into LMs to enhance their representations with structured, human-curated knowledge and improve perplexity, fact recall and downstream performance on various tasks. \citet{zhang-etal-2019-ernie,he-etal-2020-bert,wang-etal-2021-kepler} combine LM training with knowledge graph embeddings. \citet{wang2021k} introduces K-Adapter for injecting knowledge into pre-trained models that adds neural adapters for each kind of knowledge domain.
~\citet{yu-etal-2022-dict} introduces Dict-BERT, which incorporates definitions of rare or infrequent words into the input sequence and further pre-trains a BERT model.

\citet{calixto2021wikipedia} introduced a multilingual Wikipedia hyperlink prediction intermediate task to improve language model pre-training. Similarly, \citet{yasunaga-etal-2022-linkbert} introduced LinkBERT which leverages links between documents, such as hyperlinks, to capture dependencies and knowledge that span across documents by placing linked documents in the same context and pre-training the LM with MLM and document relation prediction. 

\section{Conclusion}
\looseness=-1
In this study, we introduce \escolmr{} as a multilingual, domain-adapted LM that has been further pre-trained on the ESCO taxonomy. We evaluated \escolmr{}, to the best of our knowledge, on the broadest evaluation set in this domain on 4 different langauges. The results showed that \escolmr{} outperformed \xlmr{} on job-related downstream tasks in 6 out of 9 datasets, particularly when the task was relevant to the ESCO taxonomy and context was important. It was found that the improvement of \escolmr{} was mainly due to its performance on shorter span lengths, demonstrating the value of pre-training on the ESCO dataset. \escolmr{} also demonstrated improved performance on both frequent surface spans and a wider range of spans. Overall, this work showed the potential of \escolmr{} as an LM for multilingual job-related tasks. We hope that it will encourage further research in this area.

\section*{Limitations}

There are several limitations to this study that should be considered. First, a key limitation is the lack of a variety of language-specific JAD. Here, we have four different languages namely EN, DA, FR, and DE. This means that our analysis is based on a limited subset of languages and may not be representative of JAD data outside of these four languages. 

In turn, the second limitation is that the ESCO taxonomy used as pre-training data only covers Europe and the datasets used in this work also covers mostly Europe. The results may not be generalizable to other regions. However, we see a slight improvement in the \textsc{Bhola} dataset, the data of which comes from Singapore, which hints that it could generalize to other cultures.

The ESCO relation prediction task aims for learning the relations between elements of the ESCO taxonomy. We acknowledge that we do not evaluate the effectiveness of the pre-training objective in relation-centered tasks. Unfortunately, to the best of our knowledge, there is no job-related dataset containing relations between skill/occupation concepts to benchmark our model on. We consider this interesting future work.

Finally, we did not conduct an ablation study on the ERP pre-training objective, i.e., which errors it makes. As the accuracy of the objective is 60\%, we are unable to determine which sampling method is detrimental to this accuracy. However, we suspect that the Linked sampling approach might be the hardest to predict correctly. For example, many occupations have a lot of necessary and optional skills, thus it is harder to determine if some skill truly belongs to a specific occupation. Nevertheless, we see that adding the ERP objective improves over regular MLM domain-adaptive pre-training.

Despite these limitations, we believe that this study provides valuable resources and insights into the use of \escolmr{} for analyzing JAD and suggests directions for future research. Future studies could address the limitations of this study by using a larger, more diverse datasets and by conducting ablation studies on the language model to better understand which parts contribute to the results.


\section*{Ethics Statement}
We also see a potential lack of language inclusiveness within our work, as we addressed in the Limitation section that ESCO mostly covers Europe (and the Arabic language). Nevertheless, we see \escolmr{} as a step towards inclusiveness, due to JAD frequently being English-only. In addition, to the best of our knowledge, ESCO itself is devoid of any gendered language, specifically, pronouns and other gender-specific terms in, e.g., occupations.  However, we acknowledge that LMs such as \escolmr{} could potentially be exploited in the process of hiring candidates for a specific job with unintended consequences (unconscious bias and dual use). There exists active research on fairer recommender systems (e.g., bias mitigation) for human resources (e.g., \citealp{mujtaba2019ethical, raghavan2020mitigating, deshpande2020mitigating, kochling2020discriminated, sanchez2020does, wilson2021building, vanimproving, arafan2022end}).

 \section*{Acknowledgements}

We thank both the NLPnorth and MaiNLP group for feedback on an earlier version of this paper. This research is supported by the Independent Research Fund Denmark (DFF) grant 9131-00019B and in parts by ERC Consolidator Grant DIALECT 101043235. 

\bibliography{anthology,custom}

\begin{thebibliography}{67}
\expandafter\ifx\csname natexlab\endcsname\relax\def\natexlab#1{#1}\fi

\bibitem[{Alsentzer et~al.(2019)Alsentzer, Murphy, Boag, Weng, Jindi, Naumann,
  and McDermott}]{alsentzer2019publicly}
Emily Alsentzer, John Murphy, William Boag, Wei-Hung Weng, Di~Jindi, Tristan
  Naumann, and Matthew McDermott. 2019.
\newblock \href {https://doi.org/10.18653/v1/W19-1909} {Publicly available
  clinical {BERT} embeddings}.
\newblock In \emph{Proceedings of the 2nd Clinical Natural Language Processing
  Workshop}, pages 72--78, Minneapolis, Minnesota, USA. Association for
  Computational Linguistics.

\bibitem[{Arafan et~al.(2022)Arafan, Graus, Santos, and
  Beauxis-Aussalet}]{arafan2022end}
Adam~Mehdi Arafan, David Graus, Fernando~P Santos, and Emma Beauxis-Aussalet.
  2022.
\newblock End-to-end bias mitigation in candidate recommender systems with
  fairness gates.

\bibitem[{Balog et~al.(2012)Balog, Fang, De~Rijke, Serdyukov, and
  Si}]{balog2012expertise}
Krisztian Balog, Yi~Fang, Maarten De~Rijke, Pavel Serdyukov, and Luo Si. 2012.
\newblock Expertise retrieval.
\newblock \emph{Foundations and Trends in Information Retrieval},
  6(2--3):127--256.

\bibitem[{Beauchemin et~al.(2022)Beauchemin, Laumonier, Le~Ster, and
  Yassine}]{beauchemin2022fijo}
David Beauchemin, Julien Laumonier, Yvan Le~Ster, and Marouane Yassine. 2022.
\newblock ``{FIJO}'': a french insurance soft skill detection dataset.
\newblock \emph{arXiv e-prints}, pages arXiv--2204.

\bibitem[{Bhola et~al.(2020)Bhola, Halder, Prasad, and
  Kan}]{bhola-etal-2020-retrieving}
Akshay Bhola, Kishaloy Halder, Animesh Prasad, and Min-Yen Kan. 2020.
\newblock \href {https://doi.org/10.18653/v1/2020.coling-main.513} {Retrieving
  skills from job descriptions: A language model based extreme multi-label
  classification framework}.
\newblock In \emph{Proceedings of the 28th International Conference on
  Computational Linguistics}, pages 5832--5842, Barcelona, Spain (Online).
  International Committee on Computational Linguistics.

\bibitem[{Brynjolfsson and McAfee(2011)}]{brynjolfsson2011race}
Erik Brynjolfsson and Andrew McAfee. 2011.
\newblock \emph{Race against the machine: How the digital revolution is
  accelerating innovation, driving productivity, and irreversibly transforming
  employment and the economy}.
\newblock Brynjolfsson and McAfee.

\bibitem[{Brynjolfsson and McAfee(2014)}]{brynjolfsson2014second}
Erik Brynjolfsson and Andrew McAfee. 2014.
\newblock \emph{The second machine age: Work, progress, and prosperity in a
  time of brilliant technologies}.
\newblock WW Norton \& Company.

\bibitem[{Buchmann et~al.(2022)Buchmann, Buchs, Busch, Clematide, Gnehm, and
  M{\"u}ller}]{buchmann2022swiss}
Marlis Buchmann, Helen Buchs, Felix Busch, Simon Clematide, Ann-Sophie Gnehm,
  and Jan M{\"u}ller. 2022.
\newblock Swiss job market monitor: A rich source of demand-side micro data of
  the labour market.
\newblock \emph{European Sociological Review}.

\bibitem[{Calixto et~al.(2021)Calixto, Raganato, and
  Pasini}]{calixto2021wikipedia}
Iacer Calixto, Alessandro Raganato, and Tommaso Pasini. 2021.
\newblock Wikipedia entities as rendezvous across languages: Grounding
  multilingual language models by predicting wikipedia hyperlinks.
\newblock In \emph{Proceedings of the 2021 Conference of the North American
  Chapter of the Association for Computational Linguistics: Human Language
  Technologies}, pages 3651--3661.

\bibitem[{Chan et~al.(2020)Chan, Schweter, and M{\"o}ller}]{chan2020german}
Branden Chan, Stefan Schweter, and Timo M{\"o}ller. 2020.
\newblock \href {https://doi.org/10.18653/v1/2020.coling-main.598}
  {{G}erman{'}s next language model}.
\newblock In \emph{Proceedings of the 28th International Conference on
  Computational Linguistics}, pages 6788--6796, Barcelona, Spain (Online).
  International Committee on Computational Linguistics.

\bibitem[{Chernova(2020)}]{chernova2020occupational}
Mariia Chernova. 2020.
\newblock Occupational skills extraction with {FinBERT}.
\newblock \emph{Master's Thesis}.

\bibitem[{Chung et~al.(2021)Chung, F{\'{e}}vry, Tsai, Johnson, and
  Ruder}]{chung2020rethinking}
Hyung~Won Chung, Thibault F{\'{e}}vry, Henry Tsai, Melvin Johnson, and
  Sebastian Ruder. 2021.
\newblock \href {https://openreview.net/forum?id=xpFFI\_NtgpW} {Rethinking
  embedding coupling in pre-trained language models}.
\newblock In \emph{9th International Conference on Learning Representations,
  {ICLR} 2021, Virtual Event, Austria, May 3-7, 2021}. OpenReview.net.

\bibitem[{Conneau et~al.(2020)Conneau, Khandelwal, Goyal, Chaudhary, Wenzek,
  Guzm{\'a}n, Grave, Ott, Zettlemoyer, and Stoyanov}]{conneau2020unsupervised}
Alexis Conneau, Kartikay Khandelwal, Naman Goyal, Vishrav Chaudhary, Guillaume
  Wenzek, Francisco Guzm{\'a}n, Edouard Grave, Myle Ott, Luke Zettlemoyer, and
  Veselin Stoyanov. 2020.
\newblock \href {https://doi.org/10.18653/v1/2020.acl-main.747} {Unsupervised
  cross-lingual representation learning at scale}.
\newblock In \emph{Proceedings of the 58th Annual Meeting of the Association
  for Computational Linguistics}, pages 8440--8451, Online. Association for
  Computational Linguistics.

\bibitem[{Decorte et~al.(2022)Decorte, Van~Hautte, Deleu, Develder, and
  Demeester}]{decorte2022design}
Jens-Joris Decorte, Jeroen Van~Hautte, Johannes Deleu, Chris Develder, and
  Thomas Demeester. 2022.
\newblock \href {https://arxiv.org/abs/2209.05987} {Design of negative sampling
  strategies for distantly supervised skill extraction}.
\newblock \emph{ArXiv preprint}, abs/2209.05987.

\bibitem[{Decorte et~al.(2021)Decorte, Van~Hautte, Demeester, and
  Develder}]{decorte2021jobbert}
Jens-Joris Decorte, Jeroen Van~Hautte, Thomas Demeester, and Chris Develder.
  2021.
\newblock \href {https://arxiv.org/abs/2109.09605} {Jobbert: Understanding job
  titles through skills}.
\newblock \emph{ArXiv preprint}, abs/2109.09605.

\bibitem[{Derczynski et~al.(2017)Derczynski, Nichols, van Erp, and
  Limsopatham}]{derczynski-etal-2017-results}
Leon Derczynski, Eric Nichols, Marieke van Erp, and Nut Limsopatham. 2017.
\newblock \href {https://doi.org/10.18653/v1/W17-4418} {Results of the
  {WNUT}2017 shared task on novel and emerging entity recognition}.
\newblock In \emph{Proceedings of the 3rd Workshop on Noisy User-generated
  Text}, pages 140--147, Copenhagen, Denmark. Association for Computational
  Linguistics.

\bibitem[{Deshpande et~al.(2020)Deshpande, Pan, and
  Foulds}]{deshpande2020mitigating}
Ketki~V Deshpande, Shimei Pan, and James~R Foulds. 2020.
\newblock Mitigating demographic bias in ai-based resume filtering.
\newblock In \emph{Adjunct publication of the 28th ACM conference on user
  modeling, adaptation and personalization}, pages 268--275.

\bibitem[{Devlin et~al.(2019)Devlin, Chang, Lee, and
  Toutanova}]{devlin2019bert}
Jacob Devlin, Ming-Wei Chang, Kenton Lee, and Kristina Toutanova. 2019.
\newblock \href {https://doi.org/10.18653/v1/N19-1423} {{BERT}: Pre-training of
  deep bidirectional transformers for language understanding}.
\newblock In \emph{Proceedings of the 2019 Conference of the North {A}merican
  Chapter of the Association for Computational Linguistics: Human Language
  Technologies, Volume 1 (Long and Short Papers)}, pages 4171--4186,
  Minneapolis, Minnesota. Association for Computational Linguistics.

\bibitem[{ESCO(2022)}]{esco-2022}
ESCO. 2022.
\newblock \href
  {https://esco.ec.europa.eu/en/about-esco/data-science-and-esco/machine-learning-assisted-mapping-multilingual-occupational-data-esco-part-1}
  {{M}achine {L}earning {A}ssisted {M}apping of {M}ultilingual {O}ccupational
  {D}ata to {ESCO} ({P}art 1)}.

\bibitem[{Gnehm et~al.(2022)Gnehm, Bühlmann, and
  Clematide}]{gnehm-bhlmann-clematide:2022:LREC}
Ann-Sophie Gnehm, Eva Bühlmann, and Simon Clematide. 2022.
\newblock \href {https://aclanthology.org/2022.lrec-1.414} {Evaluation of
  transfer learning and domain adaptation for analyzing german-speaking job
  advertisements}.
\newblock In \emph{Proceedings of the Language Resources and Evaluation
  Conference}, pages 3892--3901, Marseille, France. European Language Resources
  Association.

\bibitem[{Gnehm and Clematide(2020)}]{gnehm2020text}
Ann-Sophie Gnehm and Simon Clematide. 2020.
\newblock \href {https://doi.org/10.18653/v1/2020.nlpcss-1.10} {Text zoning and
  classification for job advertisements in {G}erman, {F}rench and {E}nglish}.
\newblock In \emph{Proceedings of the Fourth Workshop on Natural Language
  Processing and Computational Social Science}, pages 83--93, Online.
  Association for Computational Linguistics.

\bibitem[{Goyal et~al.(2023)Goyal, Kalra, Sharma, Mutharaju, Sachdeva, and
  Kumaraguru}]{goyal-etal-2023-jobxmlc}
Nidhi Goyal, Jushaan Kalra, Charu Sharma, Raghava Mutharaju, Niharika Sachdeva,
  and Ponnurangam Kumaraguru. 2023.
\newblock \href {https://aclanthology.org/2023.findings-eacl.163} {{J}ob{XMLC}:
  {EX}treme multi-label classification of job skills with graph neural
  networks}.
\newblock In \emph{Findings of the Association for Computational Linguistics:
  EACL 2023}, pages 2181--2191, Dubrovnik, Croatia. Association for
  Computational Linguistics.

\bibitem[{Green et~al.(2022)Green, Maynard, and
  Lin}]{green-maynard-lin:2022:LREC}
Thomas Green, Diana Maynard, and Chenghua Lin. 2022.
\newblock \href {https://aclanthology.org/2022.lrec-1.128} {Development of a
  benchmark corpus to support entity recognition in job descriptions}.
\newblock In \emph{Proceedings of the Language Resources and Evaluation
  Conference}, pages 1201--1208, Marseille, France. European Language Resources
  Association.

\bibitem[{Gururangan et~al.(2020)Gururangan, Marasovi{\'c}, Swayamdipta, Lo,
  Beltagy, Downey, and Smith}]{gururangan2020don}
Suchin Gururangan, Ana Marasovi{\'c}, Swabha Swayamdipta, Kyle Lo, Iz~Beltagy,
  Doug Downey, and Noah~A. Smith. 2020.
\newblock \href {https://doi.org/10.18653/v1/2020.acl-main.740} {Don{'}t stop
  pretraining: Adapt language models to domains and tasks}.
\newblock In \emph{Proceedings of the 58th Annual Meeting of the Association
  for Computational Linguistics}, pages 8342--8360, Online. Association for
  Computational Linguistics.

\bibitem[{Han and Eisenstein(2019)}]{han-eisenstein-2019-unsupervised}
Xiaochuang Han and Jacob Eisenstein. 2019.
\newblock \href {https://doi.org/10.18653/v1/D19-1433} {Unsupervised domain
  adaptation of contextualized embeddings for sequence labeling}.
\newblock In \emph{Proceedings of the 2019 Conference on Empirical Methods in
  Natural Language Processing and the 9th International Joint Conference on
  Natural Language Processing (EMNLP-IJCNLP)}, pages 4238--4248, Hong Kong,
  China. Association for Computational Linguistics.

\bibitem[{He et~al.(2020)He, Zhou, Xiao, Jiang, Liu, Yuan, and
  Xu}]{he-etal-2020-bert}
Bin He, Di~Zhou, Jinghui Xiao, Xin Jiang, Qun Liu, Nicholas~Jing Yuan, and Tong
  Xu. 2020.
\newblock \href {https://doi.org/10.18653/v1/2020.findings-emnlp.207}
  {{BERT}-{MK}: Integrating graph contextualized knowledge into pre-trained
  language models}.
\newblock In \emph{Findings of the Association for Computational Linguistics:
  EMNLP 2020}, pages 2281--2290, Online. Association for Computational
  Linguistics.

\bibitem[{Hochreiter et~al.(1997)Hochreiter, Schmidhuber, and
  Elvezia}]{hochreiter1997long}
Sepp Hochreiter, J\"{u}rgen Schmidhuber, and Corso Elvezia. 1997.
\newblock Long short-term memory.
\newblock \emph{Neural Computation}, 9(8):1735--1780.

\bibitem[{Hu et~al.(2020)Hu, Ruder, Siddhant, Neubig, Firat, and
  Johnson}]{hu2020xtreme}
Junjie Hu, Sebastian Ruder, Aditya Siddhant, Graham Neubig, Orhan Firat, and
  Melvin Johnson. 2020.
\newblock Xtreme: A massively multilingual multi-task benchmark for evaluating
  cross-lingual generalisation.
\newblock In \emph{International Conference on Machine Learning}, pages
  4411--4421. PMLR.

\bibitem[{Javed et~al.(2015)Javed, Luo, McNair, Jacob, Zhao, and
  Kang}]{javed2015carotene}
Faizan Javed, Qinlong Luo, Matt McNair, Ferosh Jacob, Meng Zhao, and Tae~Seung
  Kang. 2015.
\newblock Carotene: A job title classification system for the online
  recruitment domain.
\newblock In \emph{2015 IEEE First International Conference on Big Data
  Computing Service and Applications}, pages 286--293. IEEE.

\bibitem[{Javed et~al.(2016)Javed, McNair, Jacob, and Zhao}]{javed2016towards}
Faizan Javed, Matt McNair, Ferosh Jacob, and Meng Zhao. 2016.
\newblock \href {https://arxiv.org/abs/1606.00917} {Towards a job title
  classification system}.
\newblock \emph{ArXiv preprint}, abs/1606.00917.

\bibitem[{Jensen et~al.(2021)Jensen, Zhang, and
  Plank}]{jensen2021identification}
Kristian~N{\o}rgaard Jensen, Mike Zhang, and Barbara Plank. 2021.
\newblock \href {https://aclanthology.org/2021.nodalida-main.21}
  {De-identification of privacy-related entities in job postings}.
\newblock In \emph{Proceedings of the 23rd Nordic Conference on Computational
  Linguistics (NoDaLiDa)}, pages 210--221, Reykjavik, Iceland (Online).
  Link{\"o}ping University Electronic Press, Sweden.

\bibitem[{Kim(2014)}]{kim-2014-convolutional}
Yoon Kim. 2014.
\newblock \href {https://doi.org/10.3115/v1/D14-1181} {Convolutional neural
  networks for sentence classification}.
\newblock In \emph{Proceedings of the 2014 Conference on Empirical Methods in
  Natural Language Processing ({EMNLP})}, pages 1746--1751, Doha, Qatar.
  Association for Computational Linguistics.

\bibitem[{Kivim{\"a}ki et~al.(2013)Kivim{\"a}ki, Panchenko, Dessy, Verdegem,
  Francq, Bersini, and Saerens}]{kivimaki-etal-2013-graph}
Ilkka Kivim{\"a}ki, Alexander Panchenko, Adrien Dessy, Dries Verdegem, Pascal
  Francq, Hugues Bersini, and Marco Saerens. 2013.
\newblock \href {https://aclanthology.org/W13-5011} {A graph-based approach to
  skill extraction from text}.
\newblock In \emph{Proceedings of {T}ext{G}raphs-8 Graph-based Methods for
  Natural Language Processing}, pages 79--87, Seattle, Washington, USA.
  Association for Computational Linguistics.

\bibitem[{K{\"o}chling and Wehner(2020)}]{kochling2020discriminated}
Alina K{\"o}chling and Marius~Claus Wehner. 2020.
\newblock Discriminated by an algorithm: a systematic review of discrimination
  and fairness by algorithmic decision-making in the context of hr recruitment
  and hr development.
\newblock \emph{Business Research}, 13(3):795--848.

\bibitem[{Lafferty et~al.(2001)Lafferty, McCallum, and
  Pereira}]{lafferty2001conditional}
John~D. Lafferty, Andrew McCallum, and Fernando C.~N. Pereira. 2001.
\newblock Conditional random fields: Probabilistic models for segmenting and
  labeling sequence data.
\newblock In \emph{Proceedings of the Eighteenth International Conference on
  Machine Learning {(ICML} 2001), Williams College, Williamstown, MA, USA, June
  28 - July 1, 2001}, pages 282--289. Morgan Kaufmann.

\bibitem[{Lauscher et~al.(2020)Lauscher, Ravishankar, Vuli{\'c}, and
  Glava{\v{s}}}]{lauscher2020zero}
Anne Lauscher, Vinit Ravishankar, Ivan Vuli{\'c}, and Goran Glava{\v{s}}. 2020.
\newblock From zero to hero: On the limitations of zero-shot language transfer
  with multilingual transformers.
\newblock In \emph{Proceedings of the 2020 Conference on Empirical Methods in
  Natural Language Processing (EMNLP)}, pages 4483--4499.

\bibitem[{le~Vrang et~al.(2014)le~Vrang, Papantoniou, Pauwels, Fannes,
  Vandensteen, and De~Smedt}]{le2014esco}
Martin le~Vrang, Agis Papantoniou, Erika Pauwels, Pieter Fannes, Dominique
  Vandensteen, and Johan De~Smedt. 2014.
\newblock Esco: Boosting job matching in europe with semantic interoperability.
\newblock \emph{Computer}, 47(10):57--64.

\bibitem[{Lee et~al.(2020)Lee, Yoon, Kim, Kim, Kim, So, and Kang}]{leebiobert}
Jinhyuk Lee, Wonjin Yoon, Sungdong Kim, Donghyeon Kim, Sunkyu Kim, Chan~Ho So,
  and Jaewoo Kang. 2020.
\newblock Biobert: a pre-trained biomedical language representation model for
  biomedical text mining.
\newblock \emph{Bioinformatics}.

\bibitem[{Liu et~al.(2017)Liu, Chang, Wu, and Yang}]{liu2017deep}
Jingzhou Liu, Wei{-}Cheng Chang, Yuexin Wu, and Yiming Yang. 2017.
\newblock \href {https://doi.org/10.1145/3077136.3080834} {Deep learning for
  extreme multi-label text classification}.
\newblock In \emph{Proceedings of the 40th International {ACM} {SIGIR}
  Conference on Research and Development in Information Retrieval, Shinjuku,
  Tokyo, Japan, August 7-11, 2017}, pages 115--124. {ACM}.

\bibitem[{Liu et~al.(2019)Liu, Ott, Goyal, Du, Joshi, Chen, Levy, Lewis,
  Zettlemoyer, and Stoyanov}]{liu2019roberta}
Yinhan Liu, Myle Ott, Naman Goyal, Jingfei Du, Mandar Joshi, Danqi Chen, Omer
  Levy, Mike Lewis, Luke Zettlemoyer, and Veselin Stoyanov. 2019.
\newblock \href {https://arxiv.org/abs/1907.11692} {Roberta: A robustly
  optimized bert pretraining approach}.
\newblock \emph{ArXiv preprint}, abs/1907.11692.

\bibitem[{Loshchilov and Hutter(2019)}]{loshchilov2017decoupled}
Ilya Loshchilov and Frank Hutter. 2019.
\newblock \href {https://openreview.net/forum?id=Bkg6RiCqY7} {Decoupled weight
  decay regularization}.
\newblock In \emph{7th International Conference on Learning Representations,
  {ICLR} 2019, New Orleans, LA, USA, May 6-9, 2019}. OpenReview.net.

\bibitem[{Martin et~al.(2020)Martin, Muller, Ortiz~Su{\'a}rez, Dupont, Romary,
  de~la Clergerie, Seddah, and Sagot}]{martin2020camembert}
Louis Martin, Benjamin Muller, Pedro~Javier Ortiz~Su{\'a}rez, Yoann Dupont,
  Laurent Romary, {\'E}ric de~la Clergerie, Djam{\'e} Seddah, and Beno{\^\i}t
  Sagot. 2020.
\newblock \href {https://doi.org/10.18653/v1/2020.acl-main.645} {{C}amem{BERT}:
  a tasty {F}rench language model}.
\newblock In \emph{Proceedings of the 58th Annual Meeting of the Association
  for Computational Linguistics}, pages 7203--7219, Online. Association for
  Computational Linguistics.

\bibitem[{Mujtaba and Mahapatra(2019)}]{mujtaba2019ethical}
Dena~F Mujtaba and Nihar~R Mahapatra. 2019.
\newblock Ethical considerations in ai-based recruitment.
\newblock In \emph{2019 IEEE International Symposium on Technology and Society
  (ISTAS)}, pages 1--7. IEEE.

\bibitem[{Nguyen et~al.(2020)Nguyen, Vu, and Tuan~Nguyen}]{nguyen2020bertweet}
Dat~Quoc Nguyen, Thanh Vu, and Anh Tuan~Nguyen. 2020.
\newblock \href {https://doi.org/10.18653/v1/2020.emnlp-demos.2} {{BERT}weet: A
  pre-trained language model for {E}nglish tweets}.
\newblock In \emph{Proceedings of the 2020 Conference on Empirical Methods in
  Natural Language Processing: System Demonstrations}, pages 9--14, Online.
  Association for Computational Linguistics.

\bibitem[{Peters et~al.(2019)Peters, Neumann, Logan, Schwartz, Joshi, Singh,
  and Smith}]{peters-etal-2019-knowledge}
Matthew~E. Peters, Mark Neumann, Robert Logan, Roy Schwartz, Vidur Joshi,
  Sameer Singh, and Noah~A. Smith. 2019.
\newblock \href {https://doi.org/10.18653/v1/D19-1005} {Knowledge enhanced
  contextual word representations}.
\newblock In \emph{Proceedings of the 2019 Conference on Empirical Methods in
  Natural Language Processing and the 9th International Joint Conference on
  Natural Language Processing (EMNLP-IJCNLP)}, pages 43--54, Hong Kong, China.
  Association for Computational Linguistics.

\bibitem[{Raghavan et~al.(2020)Raghavan, Barocas, Kleinberg, and
  Levy}]{raghavan2020mitigating}
Manish Raghavan, Solon Barocas, Jon Kleinberg, and Karen Levy. 2020.
\newblock Mitigating bias in algorithmic hiring: Evaluating claims and
  practices.
\newblock In \emph{Proceedings of the 2020 conference on fairness,
  accountability, and transparency}, pages 469--481.

\bibitem[{S{\'a}nchez-Monedero et~al.(2020)S{\'a}nchez-Monedero, Dencik, and
  Edwards}]{sanchez2020does}
Javier S{\'a}nchez-Monedero, Lina Dencik, and Lilian Edwards. 2020.
\newblock What does it mean to'solve'the problem of discrimination in hiring?
  social, technical and legal perspectives from the uk on automated hiring
  systems.
\newblock In \emph{Proceedings of the 2020 conference on fairness,
  accountability, and transparency}, pages 458--468.

\bibitem[{Sayfullina et~al.(2018)Sayfullina, Malmi, and
  Kannala}]{sayfullina2018learning}
Luiza Sayfullina, Eric Malmi, and Juho Kannala. 2018.
\newblock Learning representations for soft skill matching.
\newblock In \emph{International Conference on Analysis of Images, Social
  Networks and Texts}, pages 141--152.

\bibitem[{Shi et~al.(2020)Shi, Yang, Guo, and He}]{shi2020salience}
Baoxu Shi, Jaewon Yang, Feng Guo, and Qi~He. 2020.
\newblock \href {https://dl.acm.org/doi/10.1145/3394486.3403338} {Salience and
  market-aware skill extraction for job targeting}.
\newblock In \emph{{KDD} '20: The 26th {ACM} {SIGKDD} Conference on Knowledge
  Discovery and Data Mining, Virtual Event, CA, USA, August 23-27, 2020}, pages
  2871--2879. {ACM}.

\bibitem[{Smith et~al.(2019)Smith, Braschler, Weiler, and
  Haberthuer}]{smith2019syntax}
Ellery Smith, Martin Braschler, Andreas Weiler, and Thomas Haberthuer. 2019.
\newblock Syntax-based skill extractor for job advertisements.
\newblock In \emph{2019 6th Swiss Conference on Data Science (SDS)}, pages
  80--81. IEEE.

\bibitem[{Stubbs and Uzuner(2015)}]{stubbs2015annotating}
Amber Stubbs and {\"O}zlem Uzuner. 2015.
\newblock Annotating longitudinal clinical narratives for de-identification:
  The 2014 i2b2/uthealth corpus.
\newblock \emph{Journal of biomedical informatics}, 58:S20--S29.

\bibitem[{Tamburri et~al.(2020)Tamburri, Van Den~Heuvel, and
  Garriga}]{tamburri2020dataops}
Damian~A Tamburri, Willem-Jan Van Den~Heuvel, and Martin Garriga. 2020.
\newblock Dataops for societal intelligence: a data pipeline for labor market
  skills extraction and matching.
\newblock In \emph{2020 IEEE 21st International Conference on Information Reuse
  and Integration for Data Science (IRI)}, pages 391--394. IEEE.

\bibitem[{van~der Goot et~al.(2021)van~der Goot, {\"U}st{\"u}n, Ramponi,
  Sharaf, and Plank}]{van-der-goot-etal-2021-massive}
Rob van~der Goot, Ahmet {\"U}st{\"u}n, Alan Ramponi, Ibrahim Sharaf, and
  Barbara Plank. 2021.
\newblock \href {https://doi.org/10.18653/v1/2021.eacl-demos.22} {Massive
  choice, ample tasks ({M}a{C}h{A}mp): A toolkit for multi-task learning in
  {NLP}}.
\newblock In \emph{Proceedings of the 16th Conference of the European Chapter
  of the Association for Computational Linguistics: System Demonstrations},
  pages 176--197, Online. Association for Computational Linguistics.

\bibitem[{van Els et~al.(2022)van Els, Graus, and
  Beauxis-Aussalet}]{vanimproving}
Sarah-Jane van Els, David Graus, and Emma Beauxis-Aussalet. 2022.
\newblock Improving fairness assessments with synthetic data: a practical use
  case with a recommender system for human resources.

\bibitem[{Vaswani et~al.(2017)Vaswani, Shazeer, Parmar, Uszkoreit, Jones,
  Gomez, Kaiser, and Polosukhin}]{vaswani2017attention}
Ashish Vaswani, Noam Shazeer, Niki Parmar, Jakob Uszkoreit, Llion Jones,
  Aidan~N. Gomez, Lukasz Kaiser, and Illia Polosukhin. 2017.
\newblock \href
  {https://proceedings.neurips.cc/paper/2017/hash/3f5ee243547dee91fbd053c1c4a845aa-Abstract.html}
  {Attention is all you need}.
\newblock In \emph{Advances in Neural Information Processing Systems 30: Annual
  Conference on Neural Information Processing Systems 2017, December 4-9, 2017,
  Long Beach, CA, {USA}}, pages 5998--6008.

\bibitem[{Wang et~al.(2021{\natexlab{a}})Wang, Tang, Duan, Wei, Huang, Ji, Cao,
  Jiang, and Zhou}]{wang2021k}
Ruize Wang, Duyu Tang, Nan Duan, Zhongyu Wei, Xuan-Jing Huang, Jianshu Ji,
  Guihong Cao, Daxin Jiang, and Ming Zhou. 2021{\natexlab{a}}.
\newblock K-adapter: Infusing knowledge into pre-trained models with adapters.
\newblock In \emph{Findings of the Association for Computational Linguistics:
  ACL-IJCNLP 2021}, pages 1405--1418.

\bibitem[{Wang et~al.(2021{\natexlab{b}})Wang, Gao, Zhu, Zhang, Liu, Li, and
  Tang}]{wang-etal-2021-kepler}
Xiaozhi Wang, Tianyu Gao, Zhaocheng Zhu, Zhengyan Zhang, Zhiyuan Liu, Juanzi
  Li, and Jian Tang. 2021{\natexlab{b}}.
\newblock \href {https://doi.org/10.1162/tacl_a_00360} {{KEPLER}: A unified
  model for knowledge embedding and pre-trained language representation}.
\newblock \emph{Transactions of the Association for Computational Linguistics},
  9:176--194.

\bibitem[{Wilson et~al.(2021)Wilson, Ghosh, Jiang, Mislove, Baker, Szary,
  Trindel, and Polli}]{wilson2021building}
Christo Wilson, Avijit Ghosh, Shan Jiang, Alan Mislove, Lewis Baker, Janelle
  Szary, Kelly Trindel, and Frida Polli. 2021.
\newblock Building and auditing fair algorithms: A case study in candidate
  screening.
\newblock In \emph{Proceedings of the 2021 ACM Conference on Fairness,
  Accountability, and Transparency}, pages 666--677.

\bibitem[{Wolf et~al.(2020)Wolf, Debut, Sanh, Chaumond, Delangue, Moi, Cistac,
  Rault, Louf, Funtowicz, Davison, Shleifer, von Platen, Ma, Jernite, Plu, Xu,
  Le~Scao, Gugger, Drame, Lhoest, and Rush}]{wolf-etal-2020-transformers}
Thomas Wolf, Lysandre Debut, Victor Sanh, Julien Chaumond, Clement Delangue,
  Anthony Moi, Pierric Cistac, Tim Rault, Remi Louf, Morgan Funtowicz, Joe
  Davison, Sam Shleifer, Patrick von Platen, Clara Ma, Yacine Jernite, Julien
  Plu, Canwen Xu, Teven Le~Scao, Sylvain Gugger, Mariama Drame, Quentin Lhoest,
  and Alexander Rush. 2020.
\newblock \href {https://doi.org/10.18653/v1/2020.emnlp-demos.6} {Transformers:
  State-of-the-art natural language processing}.
\newblock In \emph{Proceedings of the 2020 Conference on Empirical Methods in
  Natural Language Processing: System Demonstrations}, pages 38--45, Online.
  Association for Computational Linguistics.

\bibitem[{Yang et~al.(2016)Yang, Yang, Dyer, He, Smola, and
  Hovy}]{yang2016hierarchical}
Zichao Yang, Diyi Yang, Chris Dyer, Xiaodong He, Alex Smola, and Eduard Hovy.
  2016.
\newblock \href {https://doi.org/10.18653/v1/N16-1174} {Hierarchical attention
  networks for document classification}.
\newblock In \emph{Proceedings of the 2016 Conference of the North {A}merican
  Chapter of the Association for Computational Linguistics: Human Language
  Technologies}, pages 1480--1489, San Diego, California. Association for
  Computational Linguistics.

\bibitem[{Yasunaga et~al.(2022)Yasunaga, Leskovec, and
  Liang}]{yasunaga-etal-2022-linkbert}
Michihiro Yasunaga, Jure Leskovec, and Percy Liang. 2022.
\newblock \href {https://doi.org/10.18653/v1/2022.acl-long.551} {{L}ink{BERT}:
  Pretraining language models with document links}.
\newblock In \emph{Proceedings of the 60th Annual Meeting of the Association
  for Computational Linguistics (Volume 1: Long Papers)}, pages 8003--8016,
  Dublin, Ireland. Association for Computational Linguistics.

\bibitem[{Yu et~al.(2022)Yu, Zhu, Fang, Yu, Wang, Xu, Zeng, and
  Jiang}]{yu-etal-2022-dict}
Wenhao Yu, Chenguang Zhu, Yuwei Fang, Donghan Yu, Shuohang Wang, Yichong Xu,
  Michael Zeng, and Meng Jiang. 2022.
\newblock \href {https://doi.org/10.18653/v1/2022.findings-acl.150}
  {Dict-{BERT}: Enhancing language model pre-training with dictionary}.
\newblock In \emph{Findings of the Association for Computational Linguistics:
  ACL 2022}, pages 1907--1918, Dublin, Ireland. Association for Computational
  Linguistics.

\bibitem[{Zhang et~al.(2022{\natexlab{a}})Zhang, Jensen, Sonniks, and
  Plank}]{zhang-etal-2022-skillspan}
Mike Zhang, Kristian Jensen, Sif Sonniks, and Barbara Plank.
  2022{\natexlab{a}}.
\newblock \href {https://doi.org/10.18653/v1/2022.naacl-main.366}
  {{S}kill{S}pan: Hard and soft skill extraction from {E}nglish job postings}.
\newblock In \emph{Proceedings of the 2022 Conference of the North American
  Chapter of the Association for Computational Linguistics: Human Language
  Technologies}, pages 4962--4984, Seattle, United States. Association for
  Computational Linguistics.

\bibitem[{Zhang et~al.(2022{\natexlab{b}})Zhang, Jensen, and
  Plank}]{zhang-jensen-plank:2022:LREC}
Mike Zhang, Kristian~N{\o}rgaard Jensen, and Barbara Plank. 2022{\natexlab{b}}.
\newblock \href {https://aclanthology.org/2022.lrec-1.46} {Kompetencer:
  Fine-grained skill classification in danish job postings via distant
  supervision and transfer learning}.
\newblock In \emph{Proceedings of the Language Resources and Evaluation
  Conference}, pages 436--447, Marseille, France. European Language Resources
  Association.

\bibitem[{Zhang et~al.(2022{\natexlab{c}})Zhang, Jensen, van~der Goot, and
  Plank}]{zhang2022skill}
Mike Zhang, Kristian~N{\o}rgaard Jensen, Rob van~der Goot, and Barbara Plank.
  2022{\natexlab{c}}.
\newblock \href {https://arxiv.org/abs/2209.08071} {Skill extraction from job
  postings using weak supervision}.
\newblock \emph{ArXiv preprint}, abs/2209.08071.

\bibitem[{Zhang et~al.(2019)Zhang, Han, Liu, Jiang, Sun, and
  Liu}]{zhang-etal-2019-ernie}
Zhengyan Zhang, Xu~Han, Zhiyuan Liu, Xin Jiang, Maosong Sun, and Qun Liu. 2019.
\newblock \href {https://doi.org/10.18653/v1/P19-1139} {{ERNIE}: Enhanced
  language representation with informative entities}.
\newblock In \emph{Proceedings of the 57th Annual Meeting of the Association
  for Computational Linguistics}, pages 1441--1451, Florence, Italy.
  Association for Computational Linguistics.

\bibitem[{Zhao et~al.(2015)Zhao, Javed, Jacob, and McNair}]{zhao2015skill}
Meng Zhao, Faizan Javed, Ferosh Jacob, and Matt McNair. 2015.
\newblock \href {http://www.aaai.org/ocs/index.php/IAAI/IAAI15/paper/view/9363}
  {{SKILL:} {A} system for skill identification and normalization}.
\newblock In \emph{Proceedings of the Twenty-Ninth {AAAI} Conference on
  Artificial Intelligence, January 25-30, 2015, Austin, Texas, {USA}}, pages
  4012--4018. {AAAI} Press.

\end{thebibliography}
\bibliographystyle{acl_natbib}

\appendix

\section{Example Extraction from ESCO}
\label{sec:appendixa}

\begin{listing*}
\begin{minted}[frame=single,
               framesep=3mm,
               linenos=true,
               xleftmargin=15pt,
               tabsize=2]{js}
{     
    "id": int,
    "esco_code": "2250.4",
    "preferred_label": "animal therapist",
    "major_group": {
	    "title": "Veterinarians",
	    "description": "Veterinarians diagnose, [...]"
    },
    "alternative_label": [
	    "animal convalescence therapist",
	    "animal rehab therapist",
	    "animal rehabilitation therapist",
	    "animal therapists",
	    "animal therapist"
    ],
    "description": "Animal therapists provide [...]",
    "essential_skills": [
	    {
	      "title": "anatomy of animals",
	      "description": "The study of animal body parts, [...]"
	    },
	    ...
    ],
    "optional_skills": [
    	{
    	  "title": "use physiotherapy for treatment of animals",
    	  "description": "Adapt human physical therapy [...]"
    	},
	    ...
	]
  
}
\end{minted}
\caption{\textbf{Example Extraction.} An example of the information that is given for ESCO code 2250.4: animal therapist. The original page can be found here: \url{http://data.europa.eu/esco/occupation/0b2d3242-22a3-4de5-bd29-efd39cdf2c31}.} 
\label{json-example}
\end{listing*}

\clearpage

\section{Data Examples}\label{data:examples}

\begin{table}[ht]
    \centering
    \begin{tabular}{|l|l|}
    \hline
    \textsc{SkillSpan}      & \cref{json-skillspan} \\
    \textsc{Sayfullina}     & \cref{json-sayfullina} \\
    \textsc{Green}          & \cref{json-green} \\
    \textsc{Bhola}          & \cref{json-bhola} \\
    \textsc{Kompetencer}    & \cref{json-kompetencer} \\
    \textsc{Fijo}           & \cref{json-fijo} \\
    \textsc{Gnehm}          & \cref{json-gnehm} \\
    \hline
    \end{tabular}
    \caption{Data example references for each dataset.}
\end{table}

\begin{listing}[t]
\begin{minted}[frame=single,
               framesep=3mm,
               linenos=true,
               xleftmargin=15pt,
               tabsize=2]{xml}
Experience	O           O
in	        O           O
working     B-Skill     O
on          I-Skill     O
a           I-Skill     O
cloud-based I-Skill     O
application I-Skill     O
running     O           O
on          O           O
Docker      O           B-Knowledge
.           O           O

A           O           O
degree      O           B-Knowledge
in          O           I-Knowledge
Computer    O           I-Knowledge
Science     O           I-Knowledge
or          O           O
related     O           O
fields      O           O
.           O           O
\end{minted}
\caption{\textbf{Data Example SkillSpan.}} 
\label{json-skillspan}
\end{listing}

\begin{listing}[t]
\begin{minted}[frame=single,
               framesep=3mm,
               linenos=true,
               xleftmargin=15pt,
               tabsize=2]{xml}
ability     O
to          O
work        B-Skill
under       I-Skill
stress      I-Skill
condition   O

due         O
to          O
the         O
dynamic     B-Skill
nature      O
of          O
the         O
group       O
environment O
,           O
the         O
ideal       O
candidate   O
will        O
\end{minted}
\caption{\textbf{Data Example Sayfullina.}} 
\label{json-sayfullina}
\end{listing}

\begin{listing}[t]
\begin{minted}[frame=single,
               framesep=3mm,
               linenos=true,
               xleftmargin=15pt,
               tabsize=2]{xml}
A               O
sound           O
understanding   O
of              O
the             O
Care            B-Skill
Standards       I-Skill
together        O
with            O
a               O
Nursing         B-Qualification
qualification   I-Qualification
and             O
current         O
NMC             B-Qualification
registration    I-Qualification
are             O
essential       O
for             O
this            O
role            O
.               O
\end{minted}
\caption{\textbf{Data Example Green.}} 
\label{json-green}
\end{listing}

\begin{listing*}[t]
\begin{minted}[frame=single,
               framesep=3mm,
               linenos=true,
               xleftmargin=15pt,
               tabsize=2]{xml}
department economics national university singapore invites applications 
teaching oriented positions level lecturer senior lecturer [...] <labels>
\end{minted}
\caption{\textbf{Data Example Bhola.}} 
\label{json-bhola}
\end{listing*}

\begin{listing*}[t]
\begin{minted}[frame=single,
               framesep=3mm,
               linenos=true,
               xleftmargin=15pt,
               tabsize=2]{xml}
<English>
team worker                         S4
passion for developing your career  S1
liaise with internal teams          S1
identify system requirements        S2
plan out our new features           S4

<Danish>
arbejde med børn i alderen ½-3 år   S3
samarbejde                          S1
fokusere på god kommunikation       S1
bidrage til at styrke fællesskabet  S1
ansvarsbevidst                      A1
lyst til et aktivt udeliv           A1
\end{minted}
\caption{\textbf{Data Example Kompetencer.}} 
\label{json-kompetencer}
\end{listing*}

\begin{listing}[t]
\begin{minted}[frame=single,
               framesep=3mm,
               linenos=true,
               xleftmargin=15pt,
               tabsize=2]{xml}
Participer      B-relationnel
au              I-relationnel
réseau          I-relationnel
téléphonique    I-relationnel
mis             O
sur             O
pied            O
lors            O
des             O
campagnes       O
d'inscription   O
pour            O
fournir         B-pensee
les             I-pensee
renseignements  I-pensee
nécessaires     I-pensee
aux             I-pensee
assurés         I-pensee
\end{minted}
\caption{\textbf{Data Example Fijo.}} 
\label{json-fijo}
\end{listing}

\begin{listing}[t]
\begin{minted}[frame=single,
               framesep=3mm,
               linenos=true,
               xleftmargin=15pt,
               tabsize=2]{xml}
in                  O
mit                 O
guten               O
EDV-Kenntnissen     B-ICT

.                   O
Es                  O
erwartet            O
Sie                 O
eine                O
interessante        O
Aufgabe             O
in                  O
einer               O
Adressverwaltung    O
(                   O
Rechenzenter        B-ICT
)                   O
\end{minted}
\caption{\textbf{Data Example Gnehm.}} 
\label{json-gnehm}
\end{listing}

\clearpage

\begin{table*}[t]
    \centering
    \resizebox{\linewidth}{!}{
    \begin{tabular}{l|l|l|l|l}
    \toprule
                            &   Learning rate                                               &  Batch size               &\texttt{max\_seq\_length}  &  Epochs               \\
                            \midrule
    \textsc{SkillSpan}      &   \{\lr{1}{4}, \lr{5}{5}, \lr{1}{5} \lr{5}{6}\}               &  \{16, 32, 64\}           & 128                       &  20                   \\
    \textsc{Kompetencer}    &   \{\lr{1}{4}, \lr{7}{5}, \lr{5}{5}, \lr{1}{5}, \lr{5}{6}\}   &  \{8, 16, 32\}            & 128                       &  20                   \\
    \textsc{Bhola}          &   \{\lr{1}{4}, \lr{7}{5}, \lr{5}{5}, \lr{1}{5}, \lr{5}{6}\}   &  \{4, 16, 32, 64, 128\}   & \{128, 256\}              &  10                   \\
    \textsc{Sayfullina}     &   \{\lr{1}{4}, \lr{5}{5}, \lr{1}{5}\}                         &  \{16, 32, 64\}           & 128                       &  10                   \\
    \textsc{Green}          &   \{\lr{1}{4}, \lr{5}{5}, \lr{1}{5}\}                         &  \{16, 32, 64\}           & 128                       &  10                   \\
    \textsc{JobStack}       &   \{\lr{1}{4}, \lr{7}{5}, \lr{5}{5}, \lr{1}{5}, \lr{5}{6}\}   &  \{16, 32, 64, 128\}      & 128                       &  20                   \\
    \textsc{Gnehm}          &   \{\lr{1}{4}, \lr{5}{5}, \lr{1}{5}\}                         &  \{16, 32, 64\}           & 128                       &  5                   \\
    \textsc{Fijo}           &   \{\lr{1}{4}, \lr{5}{5}, \lr{1}{5}\}                         &  \{8, 16, 32, 64\}        & 128                       &  10                   \\
    \bottomrule
    \end{tabular}}  
    \caption{\textbf{Hyperparameter Sweep for Fine-tuning.} We show a hyperparameter sweep for fine-tuning all models. Learning rate differs for both \xlmr{} and \escolmr{}, where \xlmr{} performs best on lower learning rate (e.g., \lr{1}{5}) and \escolmr{} on a bit of a higher learning rate (e.g., \lr{5}{5}). A batch size of 32 works best for all models. The max sequence length is usually the same, except for \textsc{Bhola} due to it containing long texts. Epochs are determined based on previous work (i.e., the relevant datasets).}      
    \label{tab:hyperparameters}
\end{table*}

\section{Fine-tuning Details}\label{app:finetuning}

For fine-tuning \xlmr{} (+ DAPT) and \escolmr{} on the downstream tasks, we use MaChAmp~\cite{van-der-goot-etal-2021-massive}. For more details we refer to their paper. We always include the original learning rate, batch size, maximum sequence length, and epochs from the respective downstream tasks in our search space (whenever applicable). Each model is trained on an NVIDIA A100 GPU with 40GBs of VRAM and an AMD Epyc 7662 CPU. The seed numbers the models are initialized with are 276800, 381552, 497646, 624189, 884832. We run all models with the maximun number of epochs indicated in~\cref{tab:hyperparameters} and select the best-performing one based on validation set performance in the downstream metric.

\end{document}